\documentclass[letterpaper, 10 pt, conference]{ieeeconf}  

\IEEEoverridecommandlockouts                              

\usepackage{graphics} 
\usepackage{epsfig} 
\usepackage{mathptmx} 
\usepackage{times} 
\usepackage{amsmath} 
\usepackage{amssymb}  
\usepackage{multirow}
\usepackage{subfig}
\usepackage{floatrow}
\usepackage[colorlinks = true,
            linkcolor = black,
            urlcolor = blue]{hyperref}
            
\title{\LARGE \bf
Sim-to-Real Transfer for Optical Tactile Sensing
}

\author{Zihan Ding$^{1}$, Nathan F. Lepora$^{2}$, and Edward Johns$^{1}$
\thanks{$^{1}$The Robot Learning Lab at Imperial College London.}%
\thanks{$^{2}$Bristol Robotics Laboratory, University of Bristol.}%
\thanks{This work was supported by the Royal Academy of Engineering.}
}

\begin{document}

\maketitle
\thispagestyle{empty}
\pagestyle{empty}

\begin{abstract}

Deep learning and reinforcement learning methods have been shown to enable learning of flexible and complex robot controllers. However, the reliance on large amounts of training data often requires data collection to be carried out in simulation, with a number of sim-to-real transfer methods being developed in recent years. In this paper, we study these techniques for tactile sensing using the TacTip optical tactile sensor, which consists of a deformable tip with a camera observing the positions of pins inside this tip. We designed a model for soft body simulation which was implemented using the Unity physics engine, and trained a neural network to predict the locations and angles of edges when in contact with the sensor. Using domain randomisation techniques for sim-to-real transfer, we show how this framework can be used to accurately predict edges with less than 1 mm prediction error in real-world testing, without any real-world data at all.

\end{abstract}


\section{Introduction}

Tactile data offers a level of sensing granularity that is not available through vision, proprioception, or force sensing, and understanding how best to use this is an important step towards dexterous robot manipulation. In recent years, data-driven methods and robot learning have offered scalable solutions to a wide range of tasks, such as object grasping \cite{johns2016deep, levine2018learning}, contact-rich control \cite{levine2016end, martin2019variable}, and dexterous manipulation \cite{andrychowicz2018learning, kumar2016optimal}. Building upon these ideas, learning with tactile data has also been studied for tasks such as object recognition \cite{lin2019learning} and edge detection \cite{lepora2019pixels}. However, whilst learning-based methods can offer flexible solutions for developing complex controllers, they rely on large datasets -- particularly those involving deep learning and reinforcement learning (RL) -- which are typically very challenging to acquire from real-world data collection.

\begin{figure}[t]
	\centering
    \includegraphics[scale=0.13]{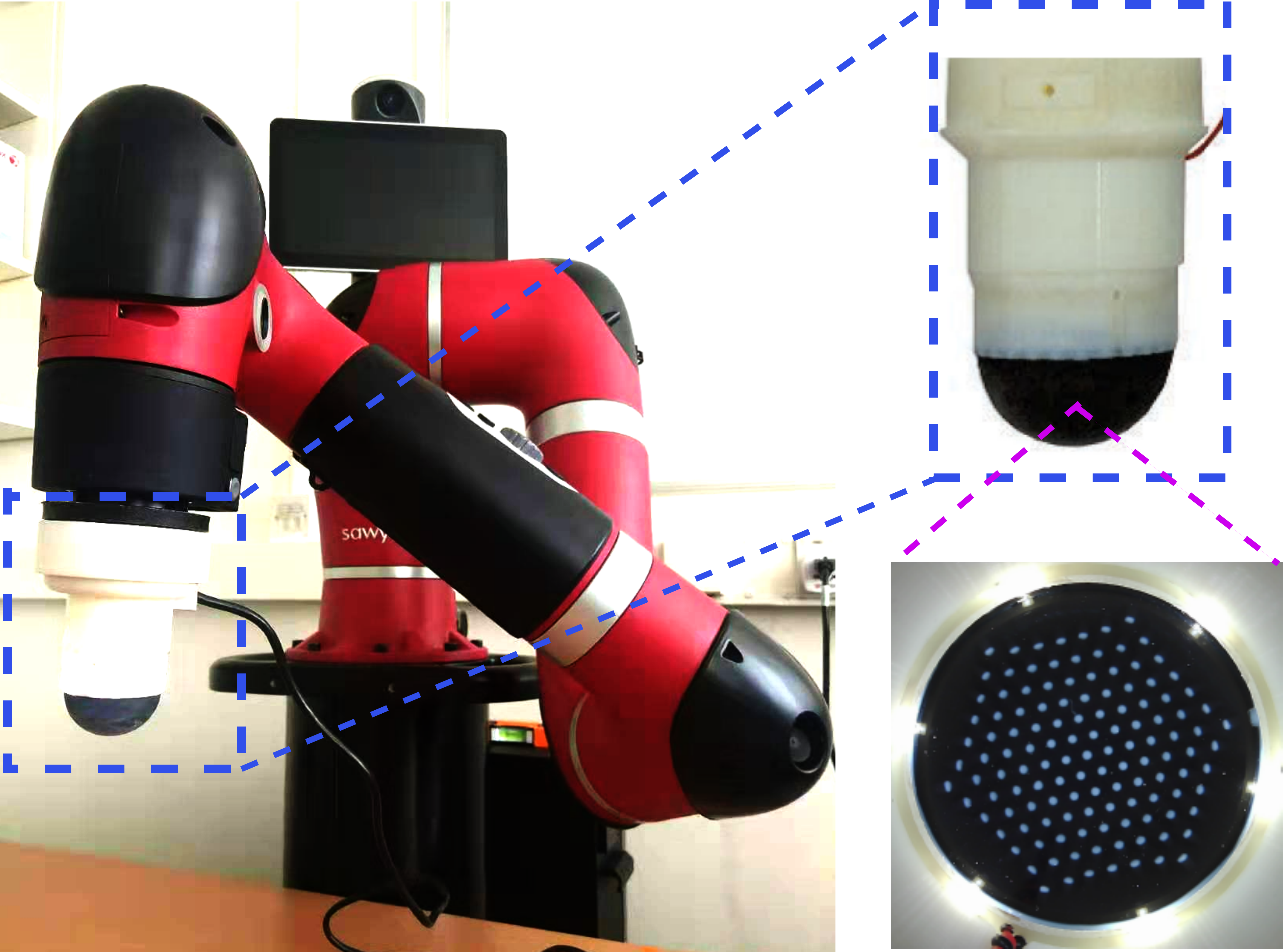}
	\caption{Our experimental setup for real world testing, with a TacTip sensor mounted onto a Sawyer robot. Here, we show the image of the sensor's pins, whose positions change as the tip deforms due to external contact.}
	\label{fig:tactip_sawyer}
\end{figure}

The field of sim-to-real transfer has made important progress in addressing this, by enabling robots to be trained in simulation with no real-world data at all \cite{johns2016deep, james2017transferring, peng2018sim}. In this paper, we study sim-to-real transfer for tactile sensing with domain randomisation \cite{tobin2017domain}, where simulation parameters are randomised to provide robustness to the difference between dynamics in simulation and reality. We use the TacTip optical sensor \cite{ward2018tactip}, which consists of two components: a deformable tip containing 127 pins on the inside surface, and a camera observing the pins as they move due to external contact. This sensor is particularly suited to sim-to-real transfer because simulation of the tip's deformation is sufficient to capture the tactile data, rather than directly simulating forces as would be necessary with sensors based on electrical signals \cite{fishel2012sensing}. In our experiments, we trained a neural network to predict positions and angles of edges using supervised learning, and tested the network in the real world by mounting the TacTip onto a Sawyer robot, as in Figure~\ref{fig:tactip_sawyer}. We studied the effect of different amounts of randomisation required during simulation, and different representations of the sensor's pin positions. Our results show that sim-to-real transfer for tactile sensing is possible, and we achieve an error of less than 1 mm in predicting the point of contact. A video of our experiments is available at \href{ http://www.robot-learning.uk/sim-to-real-tactile-icra-2020}{www.robot-learning.uk/sim-to-real-tactile-icra-2020}.

\section{Related Work}

Present work on tactile sensing for robotic control can be divided into two main categories: (1) methods which use tactile signals for basic tasks such as object identification, slip detection, and edge following \cite{lin2019learning,lepora2019pixels,lepora2017exploratory, dong2017improved,hogan2018tactile,zhang2018fingervision,garcia2019tactilegcn}; (2) methods which use tactile signals as input to reinforcement learning (RL) algorithms, for learning control policies \cite{van2016stable,huang2019learning,van2015learning}. However, these methods primarily operate with only real-world data, and as such, our work offers the potential to scale these methods to new levels of complexity via large-scale training in simulation.

For methods in the first category, Lepora et al.~\cite{lepora2017exploratory} classified edge position and angle with a probabilistic classifier, to control the motion of a robot exploring complex 2D contours. This as later extended with the use of convolutional neural networks (CNNs) for processing the tactile images~\cite{lepora2019pixels}. Lin et al.~\cite{lin2019learning} used CNNs to identify different objects from image-based tactile signals using the GelSight sensor. Hogan et al.~\cite{hogan2018tactile} used CNNs and the GelSlim sensor to predict a grasping gesture for regrasping. Zhang et al.~\cite{zhang2018fingervision} applied a convolutional long-short-term-memory (LSTM) network for slip detection using the FingerVision sensor. Garcia et al.~\cite{garcia2019tactilegcn} applied a graph convolutional network for predicting the grasp stability using the BioTac sensor.

For the second category, Hoof et al~\cite{van2016stable} used auto-encoders to process tactile and visual data for input to an RL algorithm. Huang et al.~\cite{huang2019learning} explored deep RL for gentle but contact-rich learning tasks, using the BioTac sensor and a Shadow robotic hand. Hoof et al.~\cite{van2015learning} studied in-hand manipulation using tactile data, with a policy based on a Gaussian process representation.

However, because real-world data collection is time-consuming and can cause sensors to degrade over time, sim-to-real approaches could be highly beneficial for tactile-based robot learning. Our work is based upon the use of domain randomisation~\cite{tobin2017domain, peng2018sim} for sim-to-real transfer, where simulation parameters are randomised to offer robustness to the difference between simulated and real-world dynamics.

Existing works on sim-to-real transfer for optical tactile sensors include those which learn simulation models with real-world data \cite{ruppel2018simulation, sferrazza2019ground}. Those which do not require real-world data, such as ours, include methods based on costly Finite-Element Analysis \cite{sferrazza2020learning}, and those which forego complex physics simulation entirely and assume a simplified model of deformations \cite{gomesgelsight}. Our work differs in that we explore the use of domain randomisation to avoid the need to manually determine optimal simulation parameters, and additionally, we propose the first work on sim-to-real transfer for the TacTip sensor \cite{ward2018tactip}.

\section{Methods}
\subsection{Physics Simulation}
\label{sec:deformation}

\begin{figure}[htbp]
	\centering
	\sidesubfloat[]{\includegraphics[width=50pt, height=50pt]{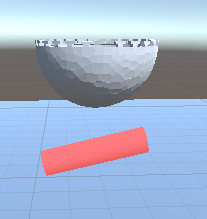}}\quad
	\sidesubfloat[]{\includegraphics[width=50pt, height=50pt]{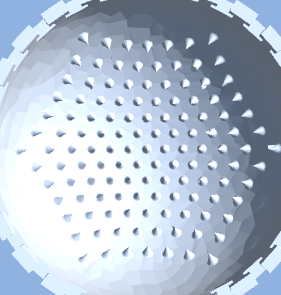}}\quad
	\sidesubfloat[]{\includegraphics[width=55pt, height=50pt]{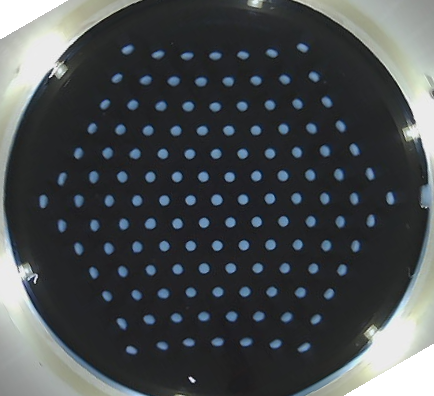}}\\
    \vspace{0.1cm}
	\sidesubfloat[]{\includegraphics[width=50pt, height=50pt]{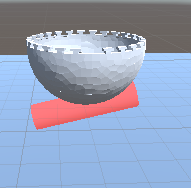}}\quad
	\sidesubfloat[]{\includegraphics[width=50pt, height=50pt]{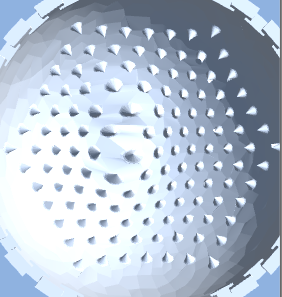}}\quad
	\sidesubfloat[]{\includegraphics[width=55pt, height=50pt]{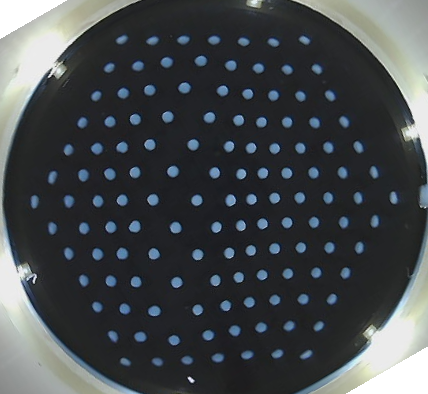}}
	\caption{The TacTip sensor's pins with no deformation (top row) and undergoing deformation due to contact (bottom row). (a), (b), (d), and (e) show the simulation, (c) and (f) show the real sensor..}
	\label{fig:deform}
\end{figure}

We used the Unity physics engine to simulate the tip's deformation. Unity allows for operations to be applied directly to vertices on an object mesh, which provides the freedom to build complex dynamic functions. We designed a model for soft body dynamics based on an approximation of real-world elastic behaviour. The membrane of the TacTip sensor in Unity before and after collision is shown in Figure~\ref{fig:deform}.
\floatsetup[figure]{style=plain,subcapbesideposition=top}

The original STL file of the TacTip sensor was first imported into Blender, and the number of vertices was reduced to enable efficient computation. Whilst Blender itself offers soft-body simulation, we found it to be very slow. We also investigated PyBullet~\cite{coumans2016pybullet} which also offers soft body simulation, but we found the simulation to be unstable in practice. Therefore, rather than using off-the-shelf soft body simulation models, we designed our own model in Unity.

We now introduce three components for approximating the elastic soft-body deformation~\cite{lee1969elastic}: the pushing force $\mathbf{f}^{push}$, the pulling force $\mathbf{f}^{pull}$, and the damping coefficient $d$ for decaying the velocity caused by pushing and pulling forces. We define the set of collision points as \{$p_c| c\in\mathcal{C}$\} and the set of vertices on the object mesh as \{$p_i| i\in\mathcal{V}$\}, where $\mathcal{C}$ and $\mathcal{V}$ are index sets correspondingly. $\mathbf{p}_{c}$ and $\mathbf{p}_{i}$ are 2-dimensional (x,y) positions for collision points and vertices, respectively, relative to the sensor's centre. We also define the vertex set of pin (centre) positions as $\mathcal{P}\subset\mathcal{V}$.

The pushing force causes the mesh's vertices to undergo acceleration, and is approximated with the following second-order inverse-distance relationship (the 1 in the denominators is for scaling and to prevent division by zero):
\begin{equation}
    \mathbf{f}^{push}_{ic} = \left(\frac{f^{push}}{1+d_{ic}^2} + \tau\frac{f^{push}}{1+d_{ic}} \right)\cdot \frac{\mathbf{d}_{ic}}{d_{ic}}, \forall i\in\mathcal{I}, \forall c\in \mathcal{C}
\end{equation}
Here, $\tau$ is a trade-off factor between linear and quadratic inverse terms, $\mathbf{d}_{ic}=\mathbf{p}_i-\mathbf{p}_c$ and $d_{ic}=|\mathbf{p}_i-\mathbf{p}_c|$ are the vector and scalar distances from the contact point $c$ to the vertex $i$, $\mathbf{f}^{push}_{ic}$ is the vector pushing force from point $c$ to vertex $i$, and $\mathcal{I}$ is the index set of vertices excluding the contact point set $\mathcal{C}$ in the overall vertex set $\mathcal{V}$ of the object: $\mathcal{I} = \{i|  p_i\neq p_c, \forall c \in \mathcal{C}\}$. Typically, very few of the overall vertices are actually in contact, and therefore $\mathcal{I}\approx \mathcal{V}$. $f^{push}$ is a constant for manually changing the magnitude of the force, which is now referred to as the \textbf{factor of pushing force}.

As a result of the pushing force, the velocity change over time $\text{d}t$ for vertex $i$ can be represented as follows:
\begin{equation}
    \Delta \mathbf{v}_i = \sum_{c\in\mathcal{C}}\mathbf{f}^{push}_{ic}\cdot \text{d} t,  \forall i\in\mathcal{I}
\end{equation}

\begin{figure}[t]
	\centering
	\includegraphics[scale=0.34]{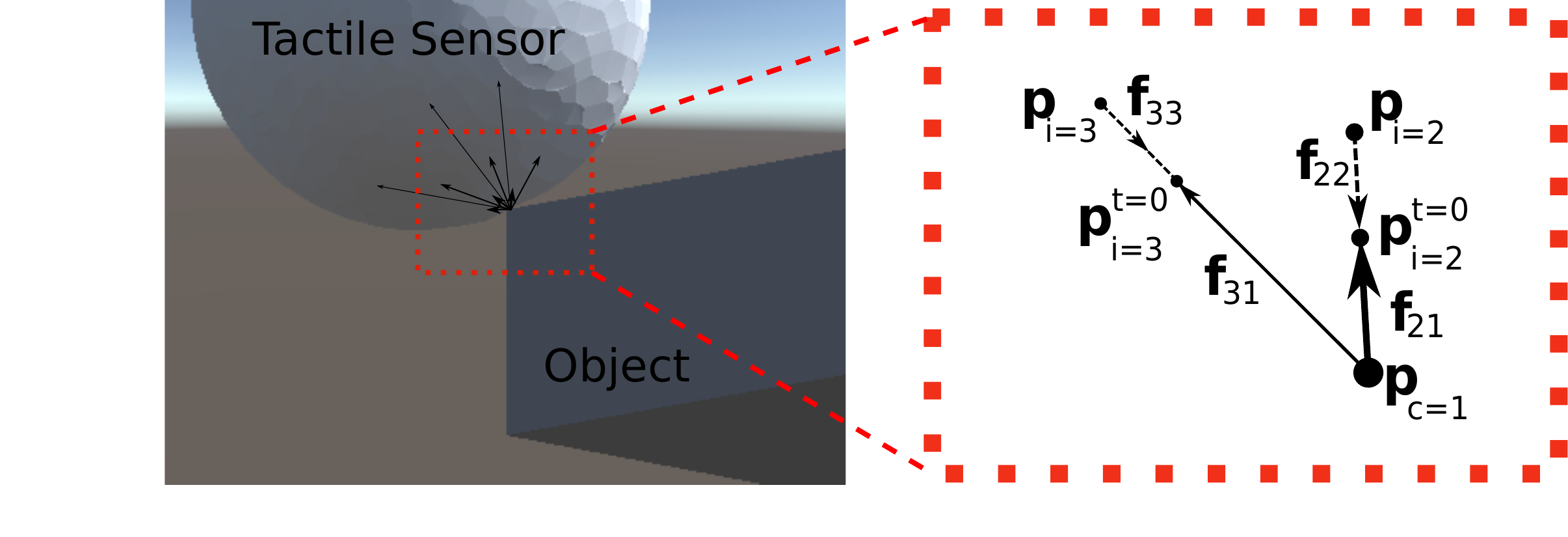}
	\caption{The pushing and pulling forces during collision of the sensor and an object. The left half shows the collision scene; the right half shows the forces during collision. $\mathbf{p}_{c=1}$ is one of the collision points, and $\mathbf{p}_{i=2}, \mathbf{p}_{i=3}$ are two vertices on the object mesh. For example, $\mathbf{f}_{21}$ is the `pushing' force from $\mathbf{p}_{c=1}$ to $\mathbf{p}_{i=2}$, and $\mathbf{f}_{22}$ is the `pulling' force in the direction opposite to the displacement of vertex $\mathbf{p}_{i=2}$.}
	\label{fig:force}
\end{figure}

Meanwhile, the pulling force causes displaced vertices to be pulled back to their original positions, due to the object's elasticity. It is defined using a first-order linear approximation of Hooke's law~\cite{rychlewski1984hooke}, in which the force magnitude is proportional to the displacement of each vertex:
\begin{equation}
    \mathbf{f}^{pull}_{ii}=f^{pull}\cdot \mathbf{d}_{ii}, \forall i\in\mathcal{I}
\end{equation}
Here, $\mathbf{d}_{ii}=\mathbf{p}^{t=0}_i-\mathbf{p}_i$ is the vector distance from the displaced vertex $i$ at $\mathbf{p}_i$ (caused by the pushing force), to its original position $\mathbf{p}^{t=0}_i$ (assuming no collision at time $t=0$), and $\mathbf{f}^{pull}_{ii}$ is the corresponding pulling force. $f^{pull}$ is a constant for manually changing the magnitude of the force, which is now referred to as the \textbf{factor of pulling force}. A schematic diagram showing the pushing and pulling forces is shown in Figure~\ref{fig:force}.

If we only consider the pushing and pulling forces, the object will continually oscillate as the pushing and pulling forces alternate in their relative magnitudes. In reality, there is an energy dissipation effect during soft body deformation, which we approximate as a \textbf{damping coefficient} $d$ on a vertex's velocity:


\begin{align}
        \Delta\mathbf{v}_i^{t\rightarrow t+\text{d}t}& = (\sum_{c\in\mathcal{C}}\mathbf{f}^{push}_{ic}+ \mathbf{f}^{pull}_{ii})\cdot \text{d} t\\
    \mathbf{v}_i^{t+\text{d}t}& = (1-d)\cdot(\mathbf{v}_i^t+\Delta\mathbf{v}_i^{t\rightarrow t+\text{d}t})
\end{align}
and the new position of vertex $i$ is:
\begin{equation}
    \mathbf{p}_i^{t+\text{d}t}=\mathbf{v}_i^t\cdot \text{d}t + \mathbf{p}_i^t
\end{equation}

Using the above model, there are three key dynamics parameters to define: (1) the factor of pushing force $f^{push}$; (2) the factor of pulling force $f^{pull}$; (3) the damping coefficient $d$. By manually tuning these three parameters, we adjusted the deformation model such that it approximately mimics soft-body deformation with visually realistic behaviour. When incorporating this model into a sim-to-real framework, these three parameters were then randomised to offer robustness to their true underlying values in real-world deformations.

\subsection{Representation of the Tactile Signal}
Using this simulation model, we can then train a neural network with supervised data, based on the sensor's 2D pin positions within the observed image of the tip's interior. In practice, we used 91 pins instead of all 127 pins, since the outermost pins were not always visible during deformation. Designing a representation for these positions is important in ensuring robust sim-to-real transfer, and we study three different representations $\mathbf{R}$ as follows:
\begin{itemize}
    \item \textbf{Pin positions} representation: the most direct way of representing TacTip deformation is directly using the 2D pin positions.  The total length of the vector is $91 \times 2 = 182$.
    
    \item \textbf{Threshold} representation: we select pins with a displacement larger than a time-dependent threshold $C^t$, and assign these pins a value of 1, and 0 otherwise. The total length of this vector is $91$. The motivation behind this representation is to remove reliance on precise modelling of pin locations, and use a coarser representation which may be more tolerant to differences between the simulator and reality. In our experiments, $C^t$ is chosen to be adaptive over time based on the current amount of deformation, by using a modified average value of pin displacements:
    \begin{equation}
        C^t=\max (\frac{1.2}{|\mathcal{P}|} \sum_{i \in \mathcal{P}} (|\mathbf{p}^t_i-\mathbf{p}^{t=0}_i|), 0.05)
    \end{equation}
    where $\mathcal{P}$ is the index set of 91 pin centres, and $\mathbf{p}^{t}_i$ is a 2-dimensional position vector of pin centre $i$ on the sensor mesh at time $t$. The 1.2 and 0.05 in above equation are constant coefficients for adopting the threshold representation, which were defined empirically. This leads to the following latent representation:
    \begin{equation}
        \mathbf{R}^t_i=H[ |\mathbf{p}_i^t-\mathbf{p}_i^{t=0}|-C^t], i\in\mathcal{P}
    \end{equation}
    where $H[.]$ is the discrete Heaviside step function. $\mathbf{R}^t$ is a list of $\mathbf{R}_i^t$ with $i\in\mathcal{P}$.
    \item \textbf{Weighted-average} representation: the weighted average position of all pins is calculated, with each pin's position weighted by its displacement, to approximate the centre point of the collision. The weighted-average representation returns the averaged 2-dimensional coordinates, together with the average displacement magnitude, and is therefore of length $3$. In practice, we found that a quadratic function of the displacements leads to a more effective weighting:
    \begin{equation}
        (x^t, y^t)=\sum_{i \in \mathcal{P}} \frac{ (\mathbf{p}_i^t-\mathbf{p}_i^{t=0})^2\mathbf{p}_i^t}
        {\sum_{i \in \mathcal{P}} (\mathbf{p}_i^t-\mathbf{p}_i^{t=0})^2}
    \end{equation}
    The average displacement magnitude is defined as:
    \begin{equation}
        \Bar{d}^t=\frac{1}{|\mathcal{P}|}\sum_{i \in \mathcal{P}} |\mathbf{p}_i^t-\mathbf{p}_i^{t=0}|
    \end{equation}
    Finally, the weighted-average representation is:
    \begin{equation}
        \mathbf{R}^t=(x^t, y^t, \Bar{d}^t)
    \end{equation}

\end{itemize}

\section{Sim-to-Sim Evaluation}
\subsection{Tasks}

\begin{figure}[t]
	\centering
	\includegraphics[scale=0.2]{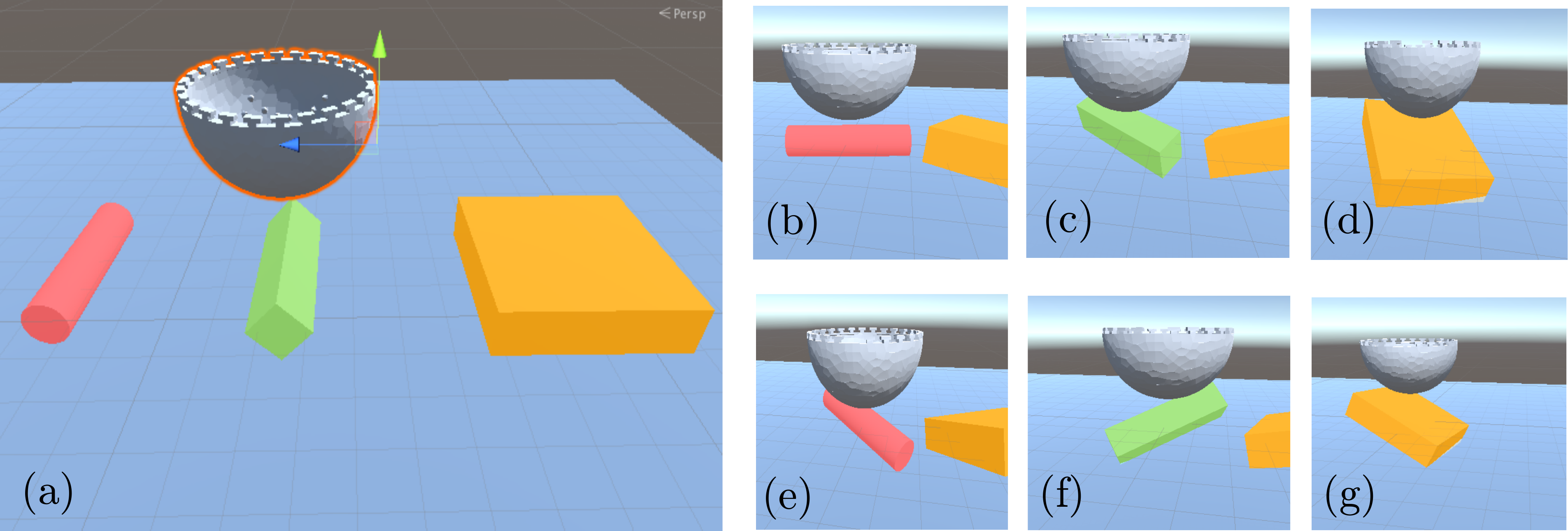}
	\caption{The three objects used during sim-to-sim evaluation. (a) shows the objects with zero rotation, (b)-(g) show random rotations. }
	\label{fig:touch_scene}
\end{figure}

Sim-to-sim transfer evaluation is a useful way to gain initial insights into the robustness of a sim-to-real transfer method, by training and testing under different simulation parameters. We studied this in a supervised learning setting, where the task is to predict (i) the object's position, (ii) the object's orientation, and (iii) the object's identity. The scene for these tests is shown in Figure~\ref{fig:touch_scene}, which contains the TacTip sensor and three different objects: a cylinder (red),  a cuboid (green), and a plane (orange).

During both training and testing, objects were randomly rotated about the $x$, $y$, and $z$ axes, with the sensor repeatedly lowered onto the object to create a set of episodes. Each episode corresponds to one \textit{tap} of the sensor onto the object, and includes several simulation steps during deformation. The objects shown in the left half of Figure \ref{fig:touch_scene} are at zero rotation about each axis. We restricted the possible rotation angles to within the range $[-30,30]$ degrees for each axis. Rotations were uniformly sampled from this range at the start of each episode, and fixed for the entire episode.

\subsection{Training}
A neural network was trained to take in the pin position representation, and predict the central position of collision, the rotation angles of the object, and the object identity. We used a four-layer fully-connected neural network, with 500 hidden units per layer and \textit{ReLU} activation~\cite{nair2010rectified}. In our experiments, we found that a convolutional neural network did not provide any improvement over a fully-connected network. The network contained two heads, with one head regressing the position and rotation with a mean squared error (MSE) loss, and the other head predicting the object identity with a \textit{Softmax} activation and MSE loss. The overall loss was the overall MSE across all predictions.

    

In order to test the effects of domain randomisation, we trained the predictor with both fixed-dynamics and random-dynamics environments, and tested the predictor with both fixed and random environments. The environments with random dynamics were achieved by randomising the pushing, pulling, and damping factors, as described previously. To generate training and testing datasets, we first empirically chose a baseline value for each parameter, which provides a good overall imitation of the real-world deformation during collisions. For the fixed-dynamics environment, we used these fixed baseline values and collected data across different positions and rotation angles of the objects relative to the sensor, as well as across different object identities. For the random-dynamics environments, we additionally randomised the values of the three dynamics parameters by applying a multiplier $\alpha$. We experimented with two different ranges for $\alpha$: $\alpha\in[0.8, 1.2]$ (\textit{randomisation factor} 0.2) and  $\alpha\in[0.5, 1.5]$ (\textit{randomisation factor} 0.5). For each task, 1000 training episodes and 100 testing episodes were collected separately, with $3\sim5$ simulation steps per episode, during which the sensor was lowered onto the object (each episode has a fixed object identity, position, and orientation, and a fixed set of dynamics parameters). Therefore, $\sim4000$ training and $\sim400$ examples were collected in simulation for each task.


\subsection{Results}

\begin{figure}[t]
	\sidesubfloat[]{\includegraphics[scale=0.12]{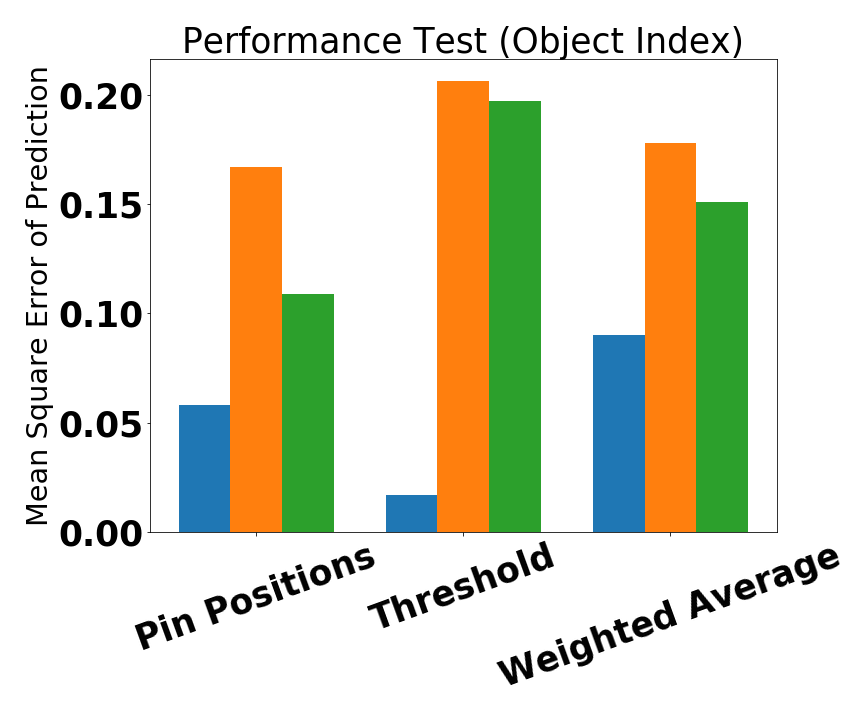}}
	\sidesubfloat[]{\includegraphics[scale=0.12]{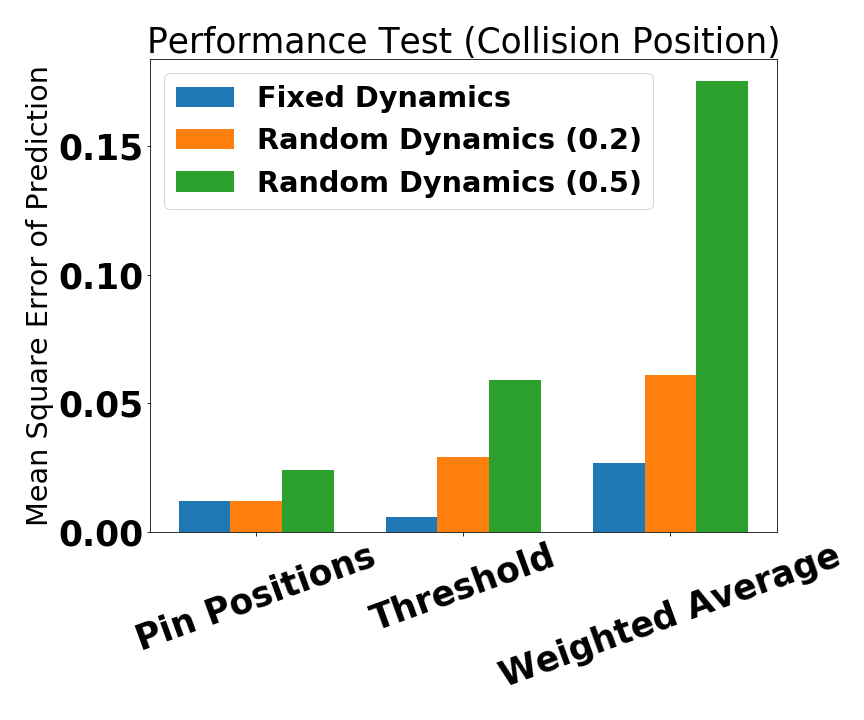}} \\
	\sidesubfloat[]{\includegraphics[scale=0.12]{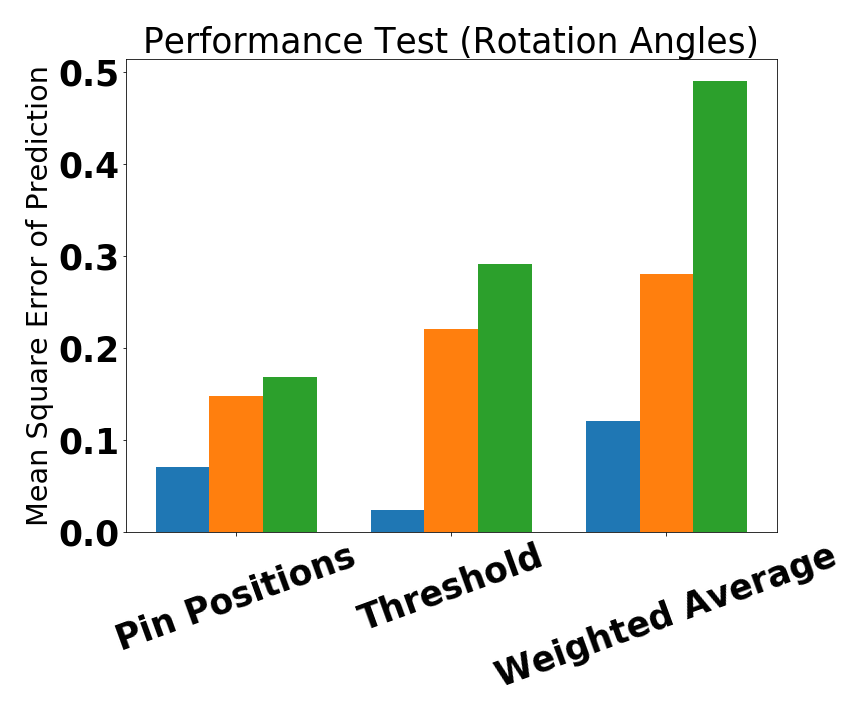}}
	\sidesubfloat[]{\includegraphics[scale=0.12]{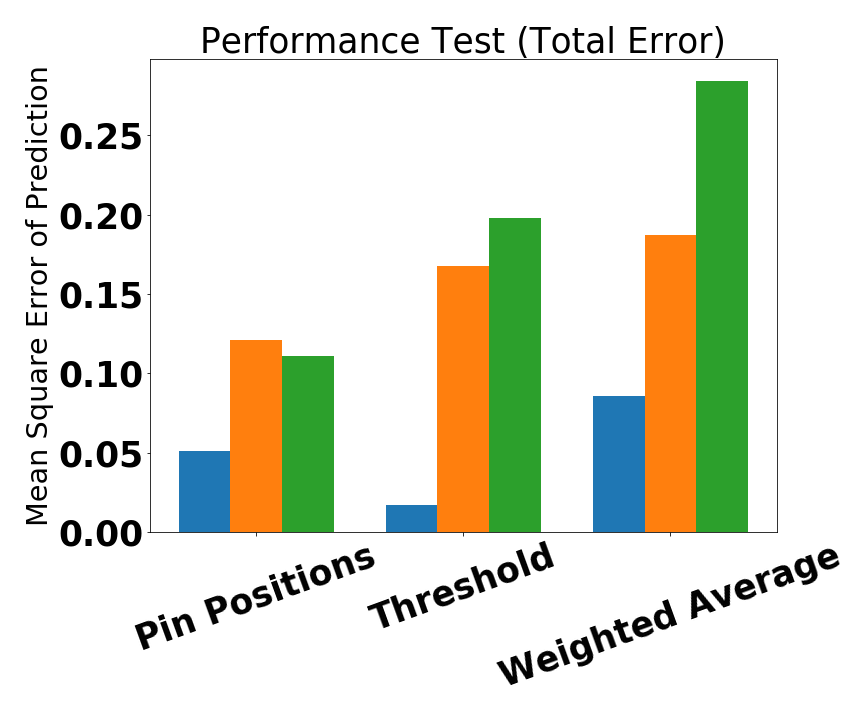}}
	\caption{Sim-to-sim comparison of different representations, with training in fixed environments but testing in randomised environments. We evaluated on three tasks: (a) predicting the object's identity, (b) predicting the central position of the collision, (c) predicting the rotation angle of the object. (d) shows the MSE of these predictions over all three tasks.}
	\label{fig:task2}
\end{figure}

Figure~\ref{fig:task2} shows test results when trained on the fixed-dynamics environment, and tested on both the fixed environment, and the random-dynamics environment with a randomisation factor of 0.2 and 0.5. We can see that the performance generally degraded as the amount of randomisation in the test environment increased, due to a greater discrepancy between training and testing simulation models. The basic pin positions representation was generally more accurate and robust than the other two representations, with the weighted-average representation performing the worst. With the threshold representation, we can see that there is overfitting to the training data since the test error under fixed dynamics is very low, but high when randomisation is introduced. This may be due to the reliance on a single threshold value, which is not flexible enough to generalise across environments. We found that the overfitting happened even when using early stopping during training.

\floatsetup[figure]{style=plain,subcapbesideposition=top}

\begin{figure}[t]
	\centering
	\sidesubfloat[]{\includegraphics[scale=0.13]{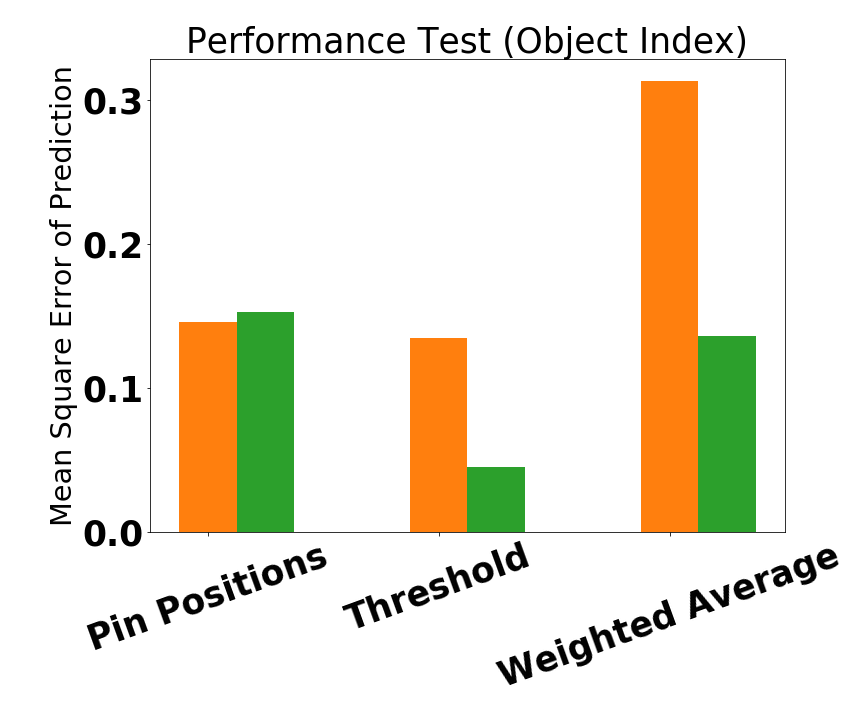}}
	\sidesubfloat[]{\includegraphics[scale=0.13]{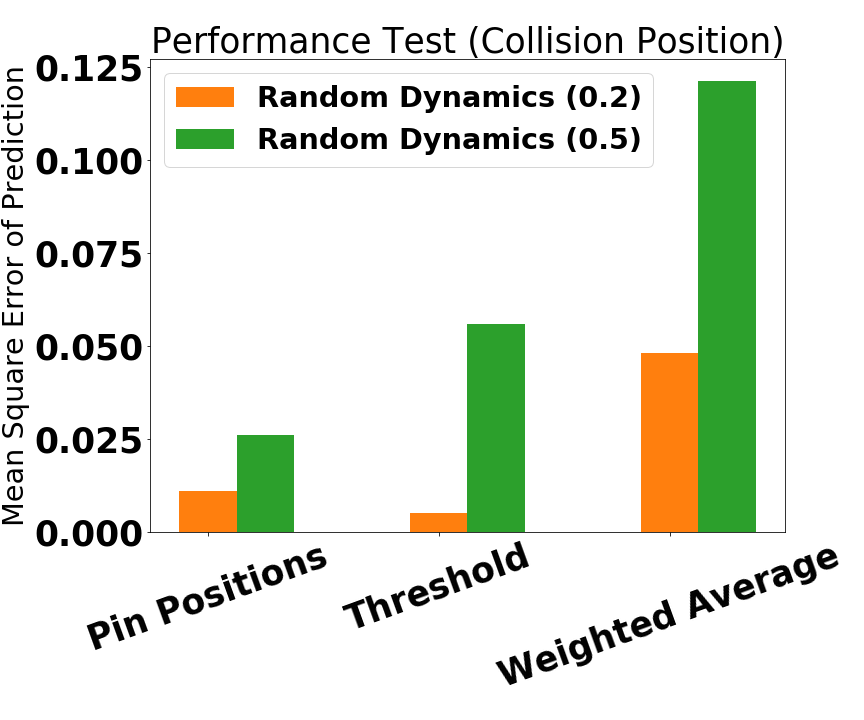}} \\
	\sidesubfloat[]{\includegraphics[scale=0.13]{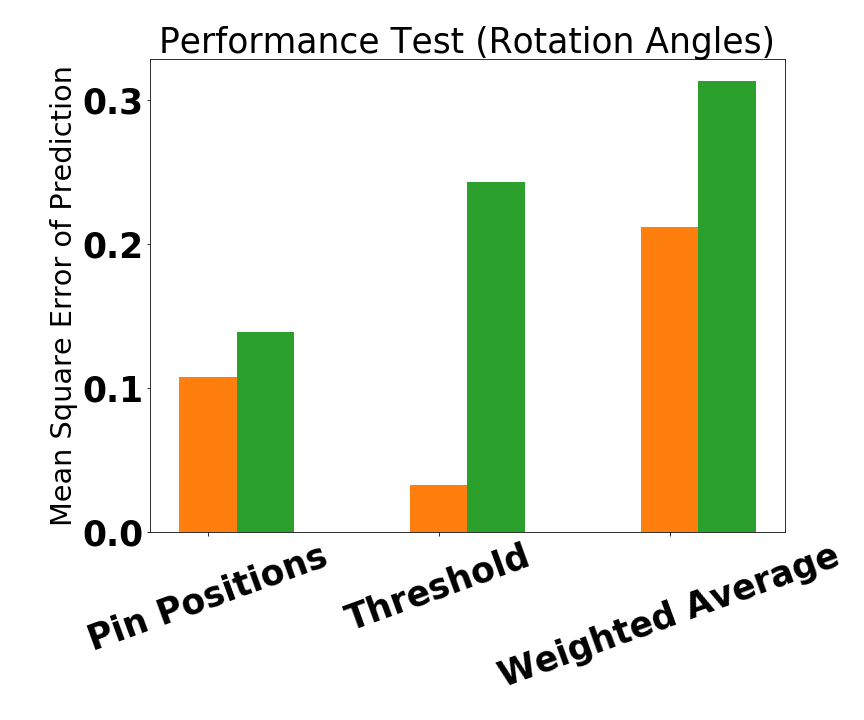}}
	\sidesubfloat[]{\includegraphics[scale=0.13]{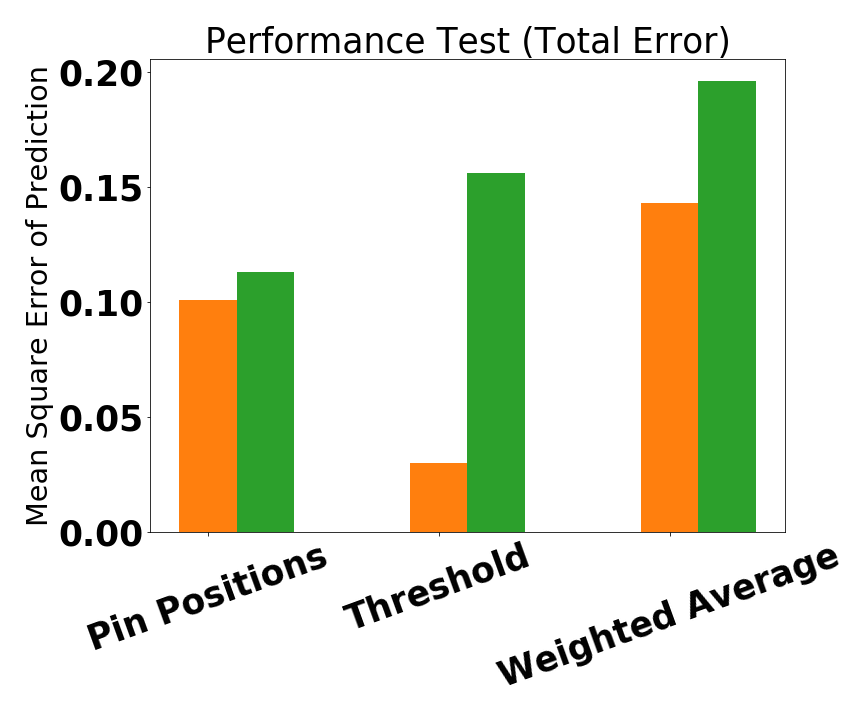}}
	\caption{Comparison of different representations, with training and testing both in randomised environments. We evaluated on three tasks: (a) predicting the object's identity, (b) predicting the central position of the collision, (c) predicting the rotation angle of the object. (d) shows the MSE of these predictions over all three tasks.}
	\label{fig:task2_random}
\end{figure}

Figure~\ref{fig:task2_random} shows test results when trained in the random-dynamics environment with a randomisation factor of 0.2, and tested in environments with factors 0.2 and 0.5. We find that training with random dynamics significantly improved the test performance when compared to training with fixed dynamics, and that using the same randomisation factor of 0.2 in both training and testing led to the best results. As before, we see that the pin positions representation performed best. In both Figures~\ref{fig:task2} and \ref{fig:task2_random}, the object index prediction shows relatively large mean prediction error. We ascribe this to the fact that the object identity is generally not predictable with a single tap of the sensor, since the objects in our tests have similar local surface geometry, even if their global geometry is distinct. 
\floatsetup[figure]{style=plain,subcapbesideposition=top}

\section{Sim-to-Real Evaluation}

We then evaluated our method with real-world experiments and sim-to-real transfer, by mounting the TacTip onto a Sawyer robot arm, as shown in Figure~\ref{fig:tactip_sawyer}. Experiments were conducted in a similar manner to the sim-to-sim setting, but with a different set of tasks and objects.

\subsection{Normalisation of Pin Positions}

First, the positions of pins when observed in the real camera need to be made consistent with those in simulation. In simulation, the pin positions are taken directly from the environment state, whereas in reality, the pin positions are estimated by use of contour localisation in the camera images. As such, normalisation is required to align the pin positions across the two domains.

To normalise the spatial unit, we first represented each pin position as the relative position from the central pin, in both simulation and reality, denoted by $\{(x^i_s, y^i_s)\}$ for simulation and $\{(x^i_r, y^i_r)\}$ for reality, where $i\in \mathcal{P}$ and $\mathcal{P}$ is the vertex set of pin centres. The normalised positions $\{(\tilde{x}^i, \tilde{y}^i)\}$ were calculated as follows:
\begin{align}
    (\tilde{x}^i, \tilde{y}^i) &= (x^i, y^i)/\frac{\bar{x}+\bar{y}}{2}, \\
    \bar{x}&=\frac{1}{|\mathcal{P}|}\sum_{i\in\mathcal{P}}|x^i|, \\
    \bar{y}&=\frac{1}{|\mathcal{P}|}\sum_{i\in\mathcal{P}}|y^i|,
\end{align}
which applies for both positions $\{(x^i_s, y^i_s)\}$ and $\{(x^i_r, y^i_r)\}$. The $\bar{x}, \bar{y}$ are the average absolute positions over all pin centres along the $x$- and $y$-axes. 

However, there were still small differences in normalised pin positions in simulation and reality after normalisation. To handle these differences, we added extra Gaussian noise (with zero mean and standard deviation of $10^{-2}$) on each pin position during training on the simulated data.


\floatsetup[figure]{style=plain,subcapbesideposition=top}
\begin{figure}[t]
	\centering
	\sidesubfloat[]{\includegraphics[height=50pt, width=60pt]{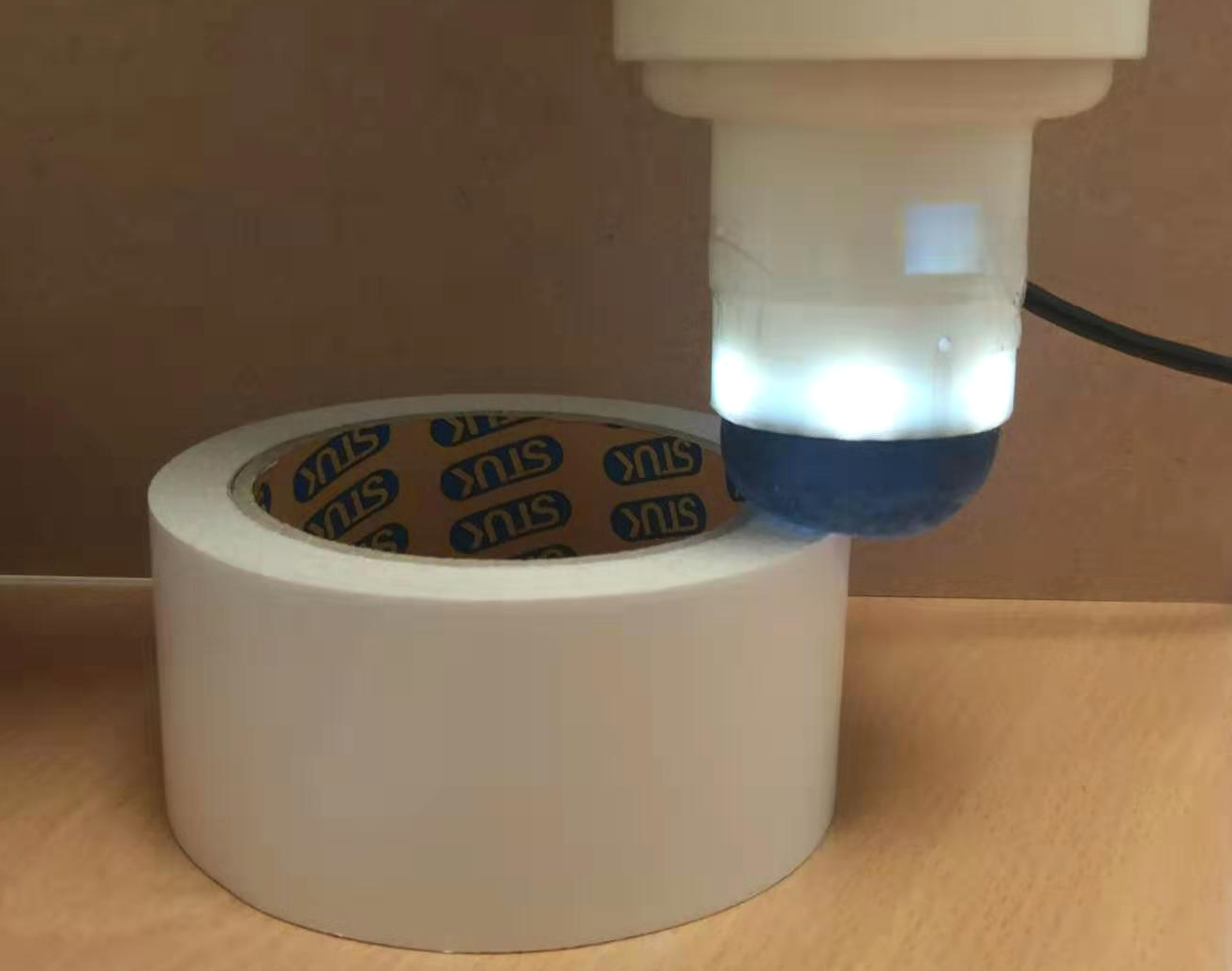}}\quad
	\sidesubfloat[]{\includegraphics[height=50pt, width=60pt]{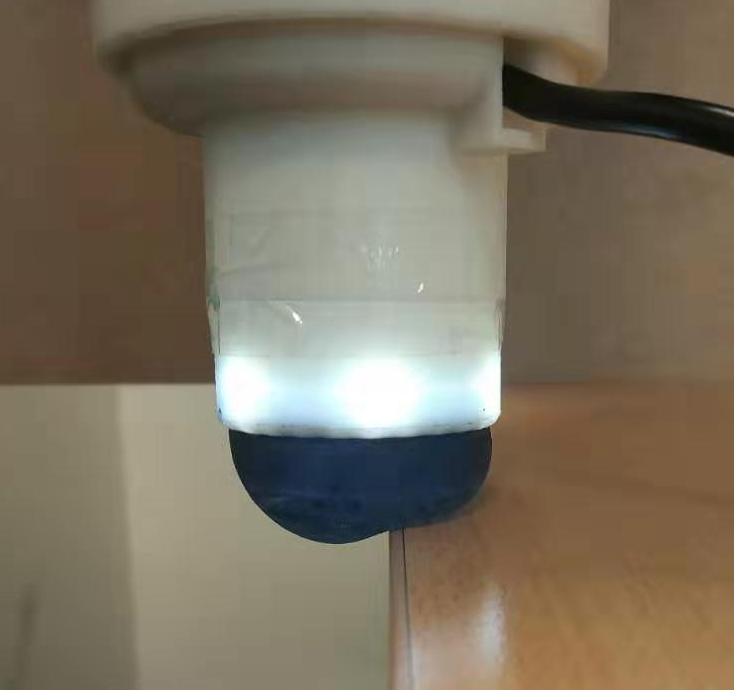}}\quad
	\sidesubfloat[]{\includegraphics[height=50pt, width=52pt]{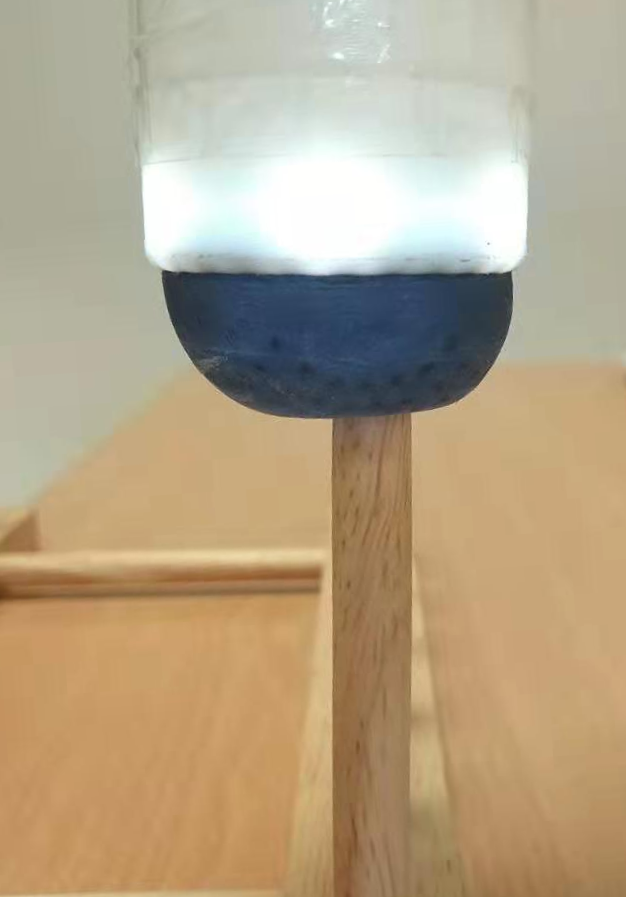}}
	\\
	\vspace{0.1cm}
	\sidesubfloat[]{\includegraphics[height=50pt, width=60pt]{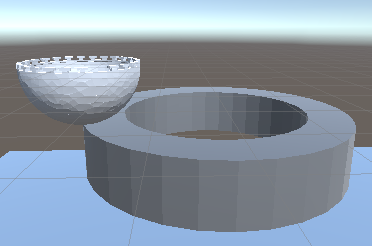}}\quad
	\sidesubfloat[]{\includegraphics[height=50pt, width=60pt]{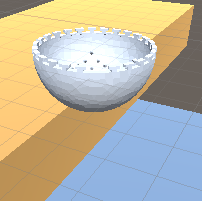}}\quad
	\sidesubfloat[]{\includegraphics[height=50pt, width=52pt]{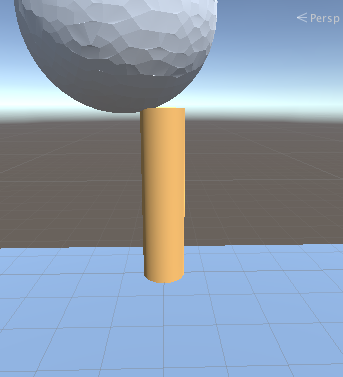}}
	
	\caption{The scenes of three tasks in both reality (a)-(c) and simulation (d)-(f). (a) and (d) are task \uppercase\expandafter{\romannumeral1}. (b) and (e) are task \uppercase\expandafter{\romannumeral2}. (c) and (f) are task \uppercase\expandafter{\romannumeral3}. }
	\label{fig:scene}
\end{figure}

\subsection{Tasks}

As illustrated in Figure~\ref{fig:scene}, we tested sim-to-real transfer on three different tasks:

\textbf{\uppercase\expandafter{\romannumeral1}. Rotation Angle Prediction}.
We tested predicting the polar coordinate angle of the sensor with respect to a circular shape, as shown in Figure~\ref{fig:scene} (a). 10000 examples were collected in simulation, containing 1000 different angles $\{\alpha=\frac{2\pi i}{1000}|i=0, 1,2,...,999\}$. In the real-world test, the sensor was moved along the circular object in a tapping manner, and tested on 12 uniformly-distributed angles around the circle $\{\alpha=\frac{2\pi i}{12}|i=0,1,2,...,11\}$. Each round therefore contained 12 taps. In both simulation and reality, small perturbations were added to the radius.

\textbf{\uppercase\expandafter{\romannumeral2}. Position Prediction on Edge}.
We tested predicting the one-dimensional position of the sensor when touching an edge, as shown in Figure~\ref{fig:scene} (b). 5000 examples were collected in simulation. In the real-world test, each round contained 10 taps on the edge. In both simulation and reality, the sensor's position was sampled uniformly from the range [-5, 5] mm between the sensor's centre and the edge.

\textbf{\uppercase\expandafter{\romannumeral3}. Position Prediction on Pole}.
We tested predicting the two-dimensional position of the sensor when touching the tip of a pole. 5000 examples were collected in simulation, recording the x- and y-coordinates of the sensor's centre relative to the pole's tip. Each round of testing contained 10 taps. During both training and testing, the sensor's position was sampled uniformly from the range [-5, 5] mm from the centre of the pole's tip.

As with the sim-to-sim experiments, data was collected in simulation across a range of randomisation factors. The total amount of real time to collect the simulated data was less than one hour for each task. For all tasks, each test was conducted for 10 rounds in the real world, and the mean absolute error (MAE) and the standard deviation of the absolute error were calculated across these rounds.

\subsection{Training}

As with the sim-to-sim experiments, training was done with a 4-layer fully-connected neural networks, with 500 hidden nodes per layer, and ReLU activations.. The rotation angle prediction used a \textit{Sigmoid} function, and the position prediction used a \textit{Tanh} function. A mean squared error loss was used for all tasks.

\subsection{Results}
\begin{figure}[htbp]
	\centering
	\includegraphics[scale=0.18]{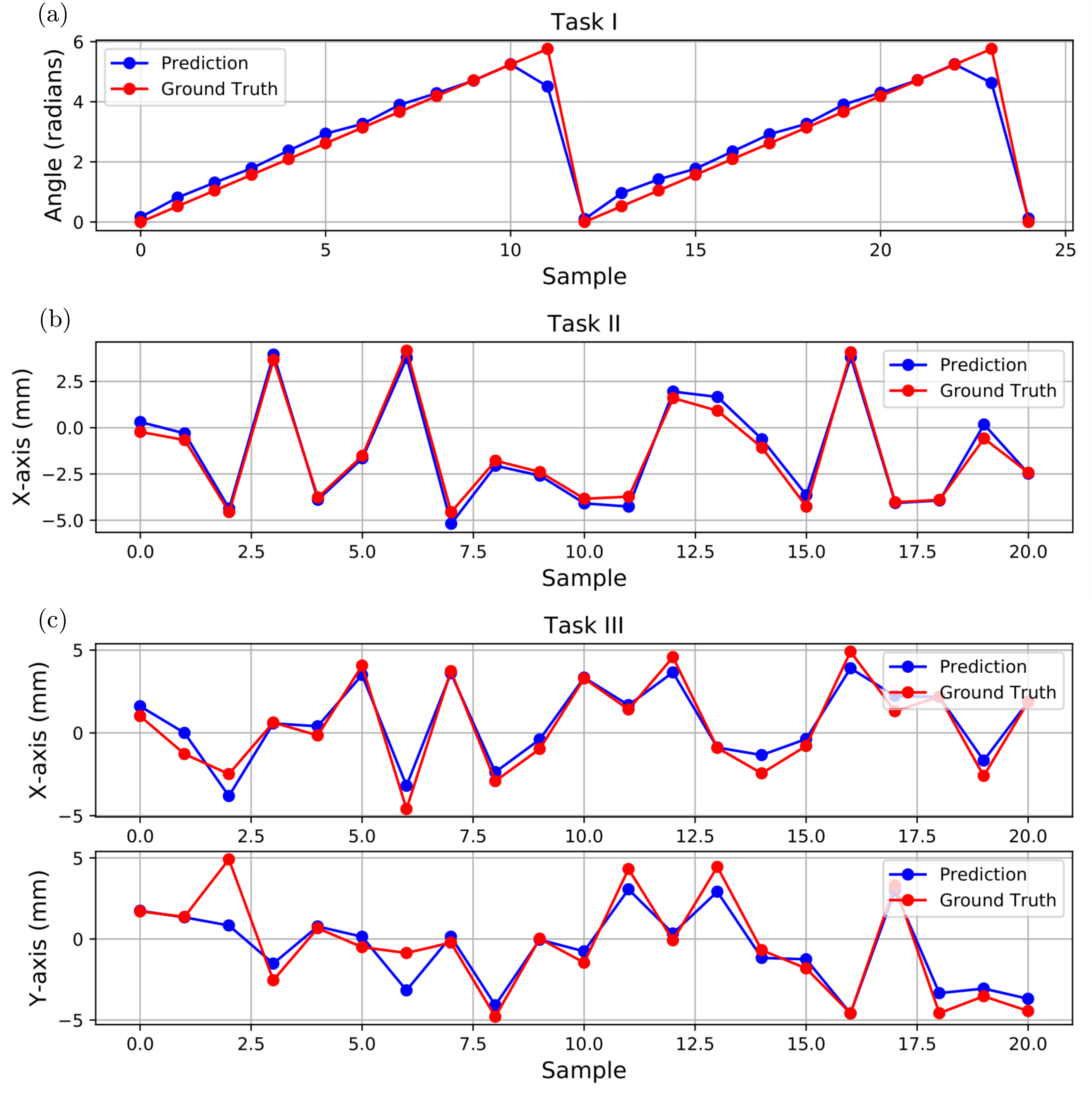}
	\caption{Prediction results for three tasks on a subset of data.}
	\label{fig:results1}
\end{figure}

\begin{figure}[htbp]
	\centering
	\sidesubfloat[]{\includegraphics[scale=0.145]{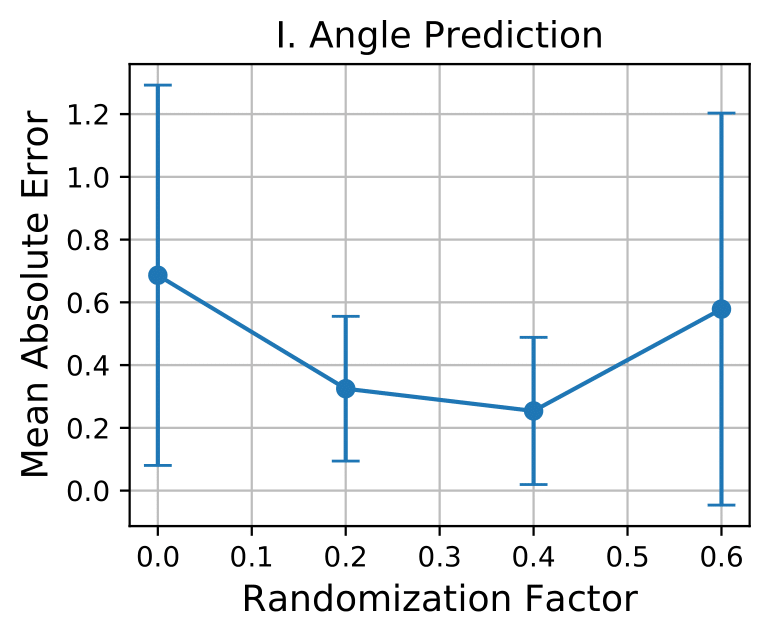}}
	\sidesubfloat[]{\includegraphics[scale=0.145]{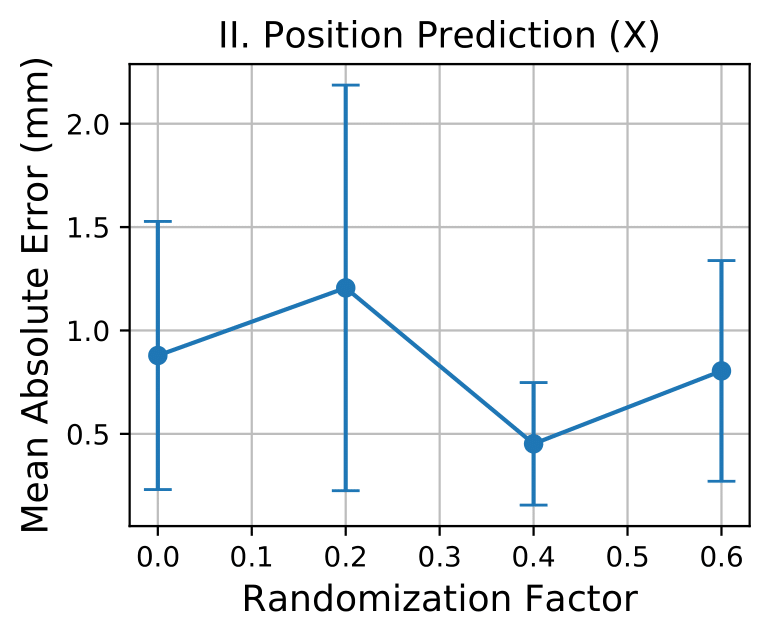}}\\
	\sidesubfloat[]{\includegraphics[scale=0.145]{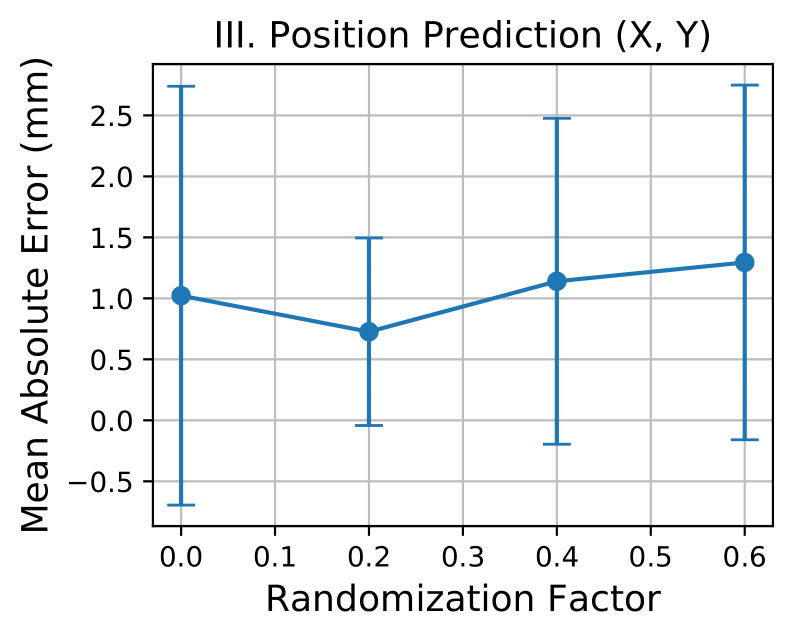}}
	\caption{Comparison of different randomisation ranges for three tasks with the pin positions representation. }
	\label{fig:results2}
\end{figure}

Results for sim-to-real transfer are displayed in Table~\ref{table}, where we also show the effect of different randomisation factors. Figure~\ref{fig:results1} (a)-(c) show the prediction performances on all tasks with a subset of data samples, using the pin positions representation, and the best randomisation range for each task (i.e. this shows our best results for each task). Figure~\ref{fig:results2} then plots the performance of the pin positions representation, as a function of the randomisation factor.

We can see three interesting outcomes from these experiments. Firstly, we show that, as with the sim-to-sim experiments, the representation which directly uses the (normalised) pin positions performs the best, and further manual engineering of the representation actually degrades performance. Secondly, we see that when using this representation, the sim-to-real transfer is robust across a wide range of parameter randomisation factors. Thirdly, we see that this method is able to achieve low errors of less than 1 mm in predicting the position of a point of contact.




\begin{table}[htbp]
\centering
\begin{tabular}{ |p{1.0cm}|p{2.0cm}|p{0.8cm}|p{0.8cm}|p{0.8cm}| p{0.8cm}| }
 \hline
 \multicolumn{6}{|c|}{Sim-to-real Performance: Mean Absolute Error} \\
 \hline
& &\multicolumn{4}{c|}{Randomisation Factor}\\
 \hline
&Representation& 0.0   & 0.2 & 0.4 & 0.6\\
\hline
 \multirow{ 3}{*}{Task \uppercase\expandafter{\romannumeral1}  } 
 & Positions & 0.686  & 0.325 & \textbf{0.254} & 0.579 \\
 &Threshold   & 3.12   & 2.71 & 3.31 & 3.20 \\
 & WA &   2.98  & 3.02   & 3.31 &  3.14\\
 \hline
 \multirow{ 3}{*}{Task \uppercase\expandafter{\romannumeral2}} 
 & Positions & 0.88  & 1.21 & \textbf{0.45} & 0.80 \\
 &Threshold   & 2.44    & 4.05 &  3.83 & 2.27 \\
 & WA &  2.24   & 2.80   & 1.41 & 1.24 \\
 \hline
 \multirow{ 3}{*}{Task \uppercase\expandafter{\romannumeral3}} 
 & Positions &  1.02 & \textbf{0.73} & 1.14 & 1.29 \\
 &Threshold   & 3.13   & 5.23 &  4.52 & 5.74\\
 & WA &   3.23  & 4.38  & 5.24 & 3.63 \\
 \hline
\end{tabular}
\caption{Real-world results studying three tasks, three pin representations, and four randomisation factors. Bold indicates the lowest error for each task.}
\label{table}
\end{table}

\floatsetup[figure]{style=plain,subcapbesideposition=top}

\section{CONCLUSIONS}
In this work, we investigated optical tactile sensing via sim-to-real transfer, and we showed this to be effective in supervised learning tasks. A soft-body simulation method with Unity was first designed to simulate the deformation of the TacTip sensor during contact. This was then used to generate training data for a number of supervised learning tasks, which involved predicting the positions, angles, and identities of objects which the sensor is in contact with. When training with a neural network, we showed that directly using the pin positions as input to the network is superior to other manually-engineered representations. Our sim-to-real results, using domain randomisation of the simulation physics parameters, show that our method can achieve less than 1 mm position prediction error, and is robust across different parameter randomisation ranges.

Whilst predicting contact points may be trivial through a manually-engineered method without requiring machine learning, the purpose of these experiments was to show that sim-to-real transfer with tactile data is possible at all. As such, this work now opens up future studies into using this for learning more complex tasks, such as those requiring reinforcement learning, where learning directly from data is required. Evaluation on more complex objects beyond simple primitive shapes, where there is greater sensitivity to the difference between simulation and reality, would also be useful in understanding real-world applicability. Additionally, limitations of the current simulation model remain to be addressed: simulation speed is currently restricted by Unity's collision detection module, and static and sliding friction effects could be incorporated during deformation, with a view towards simulating more complex object interactions.





\bibliographystyle{unsrt} 
\bibliography{IEEEexample.bib}

\end{document}